\documentclass{article}

\usepackage{microtype}
\usepackage{graphicx}

\usepackage{booktabs} 

\usepackage{hyperref}

\usepackage{amsfonts}
\usepackage{nicefrac}
\usepackage{xcolor}         
\usepackage{color}
\usepackage{colortbl}
\usepackage{soul}
\usepackage{graphicx}
\usepackage{subfig}
\usepackage{amsmath}
\usepackage{amsthm}
\usepackage{xspace}
\usepackage{soul}
\usepackage{enumitem}
\usepackage[normalem]{ulem}

\usepackage[accepted]{icml2025}

\usepackage{amsmath}
\usepackage{amssymb}
\usepackage{mathtools}
\usepackage{amsthm}

\usepackage[capitalize,noabbrev]{cleveref}

\theoremstyle{plain}

\theoremstyle{definition}

\theoremstyle{remark}

\usepackage[textsize=tiny]{todonotes}

\icmltitlerunning{Semantic-Clipping: Efficient Vision-Language Modeling
with Semantic-Guidedd Visual Selection}

\begin{document}

\twocolumn[
\icmltitle{Semantic-Clipping: Efficient Vision-Language Modeling\\ with Semantic-Guidedd Visual Selection}



\icmlsetsymbol{equal}{*}

\begin{icmlauthorlist}
\icmlauthor{Bangzheng Li}{ucd}
\icmlauthor{Fei Wang}{usc}
\icmlauthor{Wenxuan Zhou}{usc}
\icmlauthor{Nan Xu}{usc}
\icmlauthor{Ben Zhou}{asu}

\icmlauthor{Sheng Zhang}{msr}
\icmlauthor{Hoifung Poon}{msr}
\icmlauthor{Muhao Chen}{ucd}
\end{icmlauthorlist}

\icmlaffiliation{ucd}{University of California, Davis}
\icmlaffiliation{usc}{University of Southern California}
\icmlaffiliation{asu}{Arizona State University}
\icmlaffiliation{msr}{Microsoft Research}

\icmlcorrespondingauthor{Bangzheng Li}{bzhli@ucdavis.edu}

\icmlkeywords{Machine Learning, ICML}

\vskip 0.3in
]

\definecolor{codegreen}{rgb}{0,0.6,0}
\definecolor{codegray}{rgb}{0.5,0.5,0.5}
\definecolor{codepurple}{rgb}{0.58,0,0.82}
\definecolor{backcolour}{rgb}{0.95,0.95,0.92}
\definecolor{darkpink}{rgb}{0.8, 0.1, 0.5}
\definecolor{almond}{rgb}{0.94, 0.87, 0.8}
\definecolor{grannysmithapple}{rgb}{0.66, 0.89, 0.63}
\definecolor{mossgreen}{rgb}{0.68, 0.87, 0.68}
\definecolor{pearl}{rgb}{0.94, 0.92, 0.84}
\definecolor{eggshell}{rgb}{0.94, 0.92, 0.84}

\crefformat{section}{\S#2#1#3}
\crefformat{subsection}{\S#2#1#3}
\crefformat{subsubsection}{\S#2#1#3}
\crefrangeformat{section}{\S#3#1#4 to~\S#5#2#6}
\crefmultiformat{section}{\S#2#1#3}{ and~\S#2#1#3}{, #2#1#3}{ and~#2#1#3}
\Crefformat{figure}{#2Fig.~#1#3}
\Crefmultiformat{figure}{Figs.~#2#1#3}{ and~#2#1#3}{, #2#1#3}{ and~#2#1#3}
\Crefformat{table}{#2Tab.~#1#3}
\Crefmultiformat{table}{Tabs.~#2#1#3}{ and~#2#1#3}{, #2#1#3}{ and~#2#1#3}
\Crefformat{appendix}{#2Appx.~\S#1#3}
\crefformat{algorithm}{Alg.~#2#1#3}
\Crefformat{equation}{#2Eq.~#1#3}
\interfootnotelinepenalty=10000



\printAffiliationsAndNotice{}  

\newcommand{\modelname}{\textsc{SemClip}\xspace}
\newcommand{\stitle}[1]{\vspace{1ex} \noindent{\bf #1}}

\begin{abstract}

Vision-Language Models (VLMs) leverage aligned visual encoders to transform images into visual tokens, allowing them to be processed similarly to text by the backbone large language model (LLM). This unified input paradigm enables VLMs to excel in vision-language tasks such as visual question answering (VQA). To improve fine-grained visual reasoning, recent advancements in vision-language modeling introduce image cropping techniques that feed all encoded sub-images into the model. However, this approach significantly increases the number of visual tokens, leading to inefficiency and potential distractions for the LLM. To address the generalization challenges of image representation in VLMs, we propose a lightweight, universal framework that seamlessly integrates with existing VLMs to enhance their ability to process fine-grained details. 

Our method leverages textual semantics to identify key visual areas, improving VQA performance without requiring any retraining of the VLM. Additionally, it incorporates textual signals into the visual encoding process, enhancing both efficiency and effectiveness. The proposed method, \modelname, strengthens the visual understanding of a 7B VLM, LLaVA-1.5 by 3.3\% on average across 7 benchmarks, and particularly by 5.3\% on the challenging detailed understanding benchmark $V^*$.
\end{abstract}
\section{Introduction}

\begin{figure}[t!]
\begin{center}
    \includegraphics[width=0.48\textwidth]{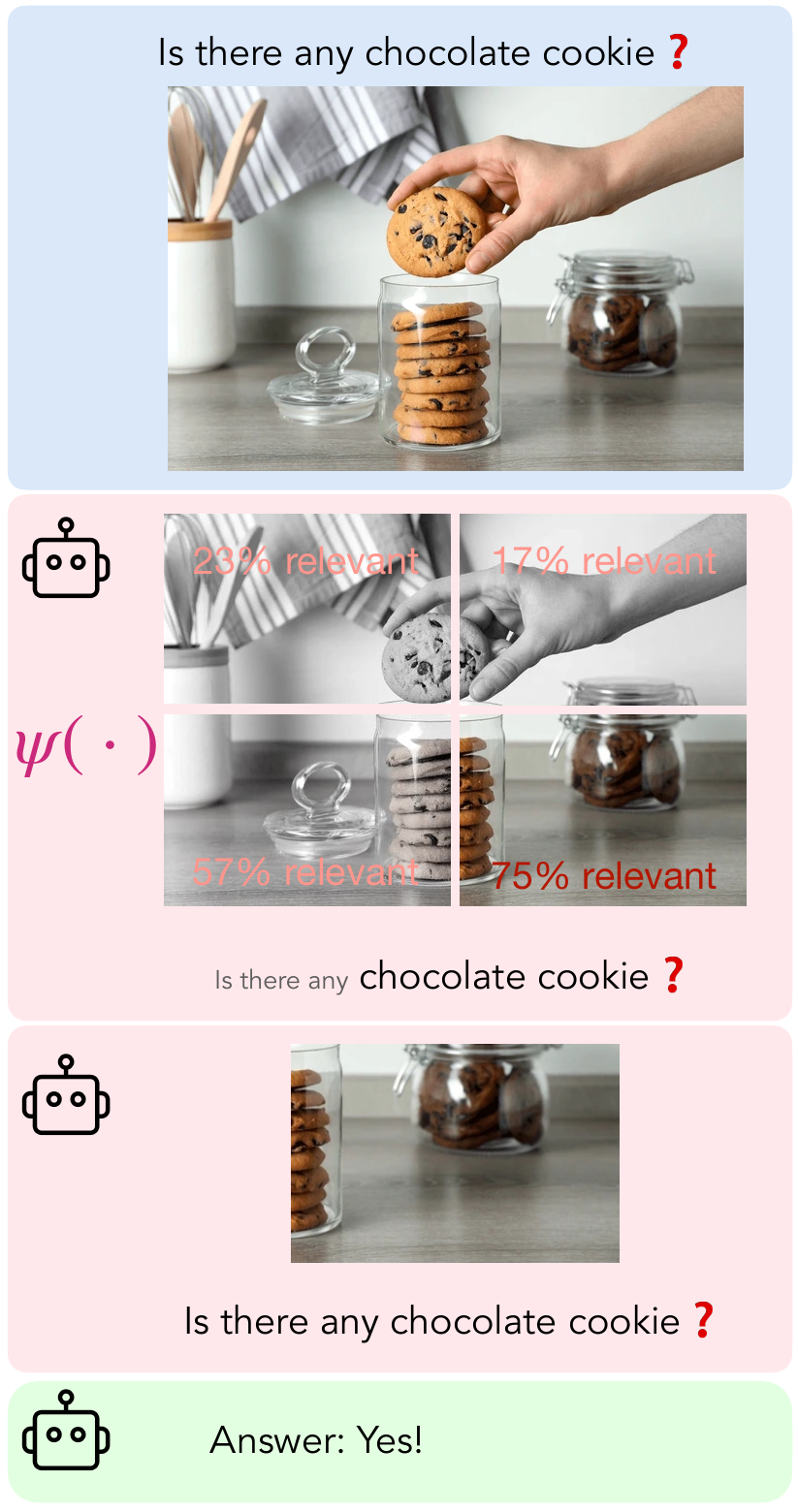}
    \caption{Paying attention to task-relevant regions within a scene is an intuitive approach to answering visual questions. Our objective is to identify the optimal task-relevance measurement \(\psi\) that selects the most pertinent sub-region of an image, enhancing visual understanding.}
    \label{fig:example}
    \vspace{-2em}
\end{center}
\end{figure}

Vision-Language Models (VLMs) have demonstrated remarkable proficiency in vision-language understanding and reasoning. Most of these models leverage aligned visual encoders to encode and map input images into a fixed set of visual tokens, which are subsequently processed as prefix tokens by a Large Language Model (LLM) in the same way as natural language tokens\cite{Flamingo,dai2023instructblip,llava10}. When handling varied image resolutions, this visual encoding process typically involves down-sizing the image to a fixed resolution and encode patches of the low-resolution image into patch-level tokens\cite{clip,vit}. However, this approach often results in shape distortions, loss of fine-grained details, and a reduction in visual signals, thereby hindering task performance, particularly for tasks requiring intricate visual comprehension\cite{llavanext,dragonfly}.

To address these limitations, recent research has highlighted the benefits of using higher-resolution encoders, demonstrating that reducing excessive downsampling improves performance across various tasks. Additionally, methods such as LLaVA-NeXT incorporate multi-crop techniques, segmenting an image into multiple parts to enable processing at or near native resolution. While these methods preserve more visual details, they also generate significantly larger number of tokens, which can strain the context limits of LLMs, increase computational costs, and potentially hinder the LLMs' reasoning performance.


To address the challenge of handling a large number of visual tokens, we introduce semantic-guided sub-image selection. Humans inherently retain task-related instructions while interacting with their environment. For instance, as illustrated in \Cref{fig:example}, when searching for a chocolate cookie, one would first identify the jar containing chocolate cookies, focus on the relevant area, and then make a decision. Similarly, VLMs must prioritize image regions that are directly relevant to the given query in order to generate accurate responses. 

To achieve this, we propose \modelname, a novel sub-image selection method designed to enhance the fine-grained visual understanding of existing VLMs without requiring additional fine-tuning. \modelname functions as a plug-and-play approach that improves high-resolution visual comprehension while minimizing the number of extra visual tokens needed. During inference, \modelname first segments the original image into multiple candidate sub-images. It then evaluates the \textbf{relevance} of each sub-image to the given textual query, selecting the most relevant one to append to the original image before feeding it into the VLM. We explore various relevance-scoring strategies, which can leverage extracted features from VLM modules, pre-trained language-image models, or efficiently fine-tuned models that require minimal training on distantly supervised data.

Our proposed approach offers several merits to VLMs. First, experimental results demonstrate that \modelname significantly enhances the visual question answering (VQA) capabilities of widely used VLMs. Second, \modelname achieves this improvement with minimal additional computational overhead, ensuring both performance gains and token efficiency. Analysis, as detailed in \Cref{ssec:visUnderstand,ssec:tokenNums}, reveals that \modelname attains comparable or superior performance on VQA benchmarks while requiring fewer computation resources compared to baseline methods. Finally, \modelname is a plug-in solution compatible with common VLMs, enabling high-resolution visual understanding without the need for model retraining.

In summary, the key contributions of this work are as follows:  
(1) We propose \modelname, a plug-and-play approach that integrates textual signals into the visual encoding process, significantly improving the visual understanding capabilities of modern VLMs.  
(2) We demonstrate the effectiveness and computational efficiency of \modelname through extensive experiments.  
(3) We train and publicly release a CLIP-based model optimized to enhance the performance of \modelname.  
(4) We provide experimental evidence showing that text-guided visual selection has a high theoretical upper bound, highlighting the substantial gap between the current visual understanding capabilities of VLMs and their full potential.
\section{\modelname}\label{sec:model}
In this section, we begin by outlining the fundamentals of the vision-language model in \Cref{ssec:preliminary}. Next, we describe the \modelname pipeline in \Cref{ssec:semclipMethod}, with a focus on its core component—relevance measurement—which is detailed in \Cref{ssec:relevancMethod}.

\begin{figure*}[t!]
\begin{center}
    \includegraphics[width=0.8\textwidth]{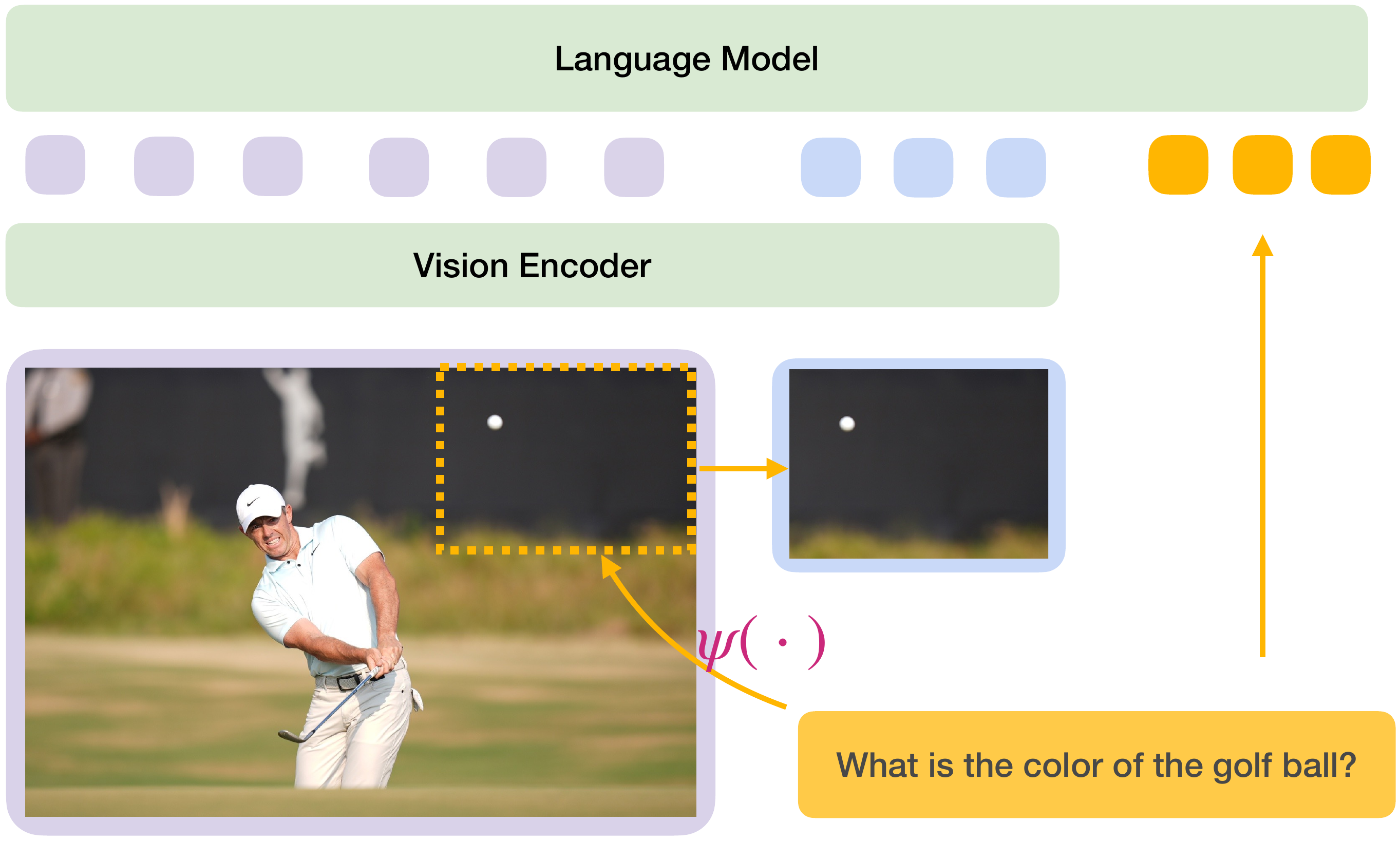}
    \caption{\modelname is a plug-and-play method that enhances a VLM through semantic-guided visual selection. Task-relevant sub-areas of the image was encoded and appended to the visual tokens of the overview image(colored in blue and purple, respectively). These visual tokens together with text embeddings of the question (colored in orange) are processed by the backbone VLM to generate a response. }
    \label{fig:pipeline}
    \vspace{-1em}
\end{center}
\end{figure*}

\subsection{Preliminaries}\label{ssec:preliminary}
Modern VLMs consist of three key components: a vision encoder \( g(\cdot) \), an LLM backbone \( f_{\phi}(\cdot) \), and a multimodal projection module \( w(\cdot) \) that bridges the two modalities. Given an input consisting of an image-question pair \((\mathbf{X}_v, \mathbf{X}_q)\), the vision encoder first processes the image \(\mathbf{X}_v\) into visual features, represented as \( g(\mathbf{X}_v) = \mathbf{Z}_v \), where \(\mathbf{Z}_v\) is a fixed-length sequence of visual feature vectors. Meanwhile, the textual question \(\mathbf{X}_q\) is encoded into language features \(\mathbf{H}_q\) through the embedding layers of the language model. Since the visual and textual features reside in different latent spaces, the multimodal projection module \( w(\cdot) \) transforms the visual features into a representation compatible with the language model’s latent space, denoted as \( w(\mathbf{Z}_v) = \mathbf{H}_v \).
This projection module can take various forms, such as an MLP~\cite{llava15,he2024efficient}, or a cross-attention module~\cite{Flamingo} etc.
Finally, the aligned visual and textual features, \(\mathbf{H}_v\) and \(\mathbf{H}_q\), are fed into the language model to generate a response: \( f_{\phi}(\mathbf{H}_v, \mathbf{H}_q) \).

\subsection{Problem Definition}
Current VLMs typically process images at a fixed resolution~\cite{llava10,chen2023shikra}. For high-resolution images, this mechanism requires either cropping or downscaling to fit the model’s input restriction, often leading to the loss of fine-grained details or distortions of images.
To mitigate this issue, recent methods~\cite{llavanext} propose to segment high-resolution images into multiple patches, encoding each independently, and using all as inputs.
Specifically, given an image-question pair \((\mathbf{X}_v, \mathbf{X}_q)\), the image \(\mathbf{X}_v\) is divided into an \( n \times n \) grid of smaller sub-images, denoted as \( \{\mathbf{X}_{v_1}, ..., \mathbf{X}_{v_{n \times n}}\} \). Each sub-image is then processed separately by the vision encoder \( g(\cdot) \) and the projection module \( w(\cdot) \), producing a set of visual token sequences \( \{\mathbf{H}_{v_1}^q, ..., \mathbf{H}_{n \times n}^q\} \).
Finally, all sub-images, along with the original image and the question, are fed into the language model to generate a response:
\[
f_{\phi}(\mathbf{H}_v, \mathbf{H}_{v_1}^q, ..., \mathbf{H}_{v_{n \times n}}^q, \mathbf{H}_q).
\]
While this approach preserves high-resolution details, it significantly increases the number of visual tokens, leading to higher computational costs.

To address these challenges, we aim to select a subset of high-resolution subimages while adhering to a restriction on the total number of tokens.
We formulate this as a visual token selection problem. Specifically, we seek a subset of high-resolution subimages of size \( k \ll n \) that maximizes model performance, as measured by evaluator $\mathcal{P}$:
\begin{align*}
    S^* = {\arg\max}_{S \subseteq \{ \mathbf{H}_{v_1}^q, ..., \mathbf{H}_{v_{n \times n}}^q \}} \mathcal{P}(f_{\phi}(\mathbf{H}_v, S, \mathbf{H}_q))\\
    \text{s.t.} \quad |S| = k
\end{align*}
Note that $k$ can be seen as a hyperparameter depending on the resource limit.
Since \( \mathcal{P} \) is typically not accessible during inference, directly optimizing for \( S^* \) is not feasible. Therefore, we need to develop an approximation method to estimate the optimal subset \( S^* \) efficiently.

\subsection{Semantic Clipping}\label{ssec:semclipMethod}
We propose to adopt a semantic-aware image clipping mechanism. Intuitively, in vision-language tasks such as VQA, not all image regions are pertinent to the given question. Including the irrelevant visual tokens not only increases computational overhead but also risks diverting the language model’s attention away from the informative sub-images.
To this end, we introduce a relevance scoring function, \( \psi \), which evaluates the relevance of each sub-image $\mathbf{X}_{v_i}$ with respect to the question \( \mathbf{X}_q \). Then, rather than processing visual features from all sub-images, we retain only the top-\( k \) most relevant sub-images
\begin{equation*}
\resizebox{0.49\textwidth}{!}{
$
S = \left\{ \mathbf{X}_{v_i} \mid i \in \arg\text{top-k} \left(  
\psi(\mathbf{X}_q, \mathbf{X}_{v_1}), \dots,  
\psi(\mathbf{X}_q, \mathbf{X}_{v_{n \times n}})  
\right) \right\}.
$}
\end{equation*}
as input to the language model.
This approach effectively reduces the number of visual tokens while ensuring that the retained visual features contribute meaningfully to the final response. 



\subsection{Relevance Measurement}\label{ssec:relevancMethod}
We explore multiple approaches to assess the relevance of a sub-image with respect to a given question.  
We first consider \textbf{semantic similarity-based methods}, which evaluate how closely a sub-image aligns with the question. The underlying assumption is that if a sub-image exhibits high similarity to the question, it is likely to contain relevant information that helps answer it. Within this category, we examine two alternatives: using the VLM itself as a semantic similarity measure, or utilizing an external pretrained similarity model.
However, relevance may extend beyond semantic similarity alone. For instance, given the question \textit{``What should be used to cut a watermelon?''}, a sub-image containing a watermelon may exhibit a high semantic similarity, yet it does not provide the necessary information to answer the question. Instead, the model should prioritize attending to a sub-image of a knife, as it directly pertains to the action described in the query.  
To capture possibly complex relationships between a question and its corresponding sub-images, we propose a \textbf{distant supervision approach} that directly trains a relevance model using distantly labeled data.  
We provide a detailed discussion of each method in the remaining of the section.

\smallskip
\noindent\textbf{VLM-based similarity \( \psi_{lm} \).}
In this approach, we utilize the VLM itself to measure the similarity between a sub-image and the given question. Each sub-image is processed independently through the vision encoder and the multimodal projection module. We extract the projected classification visual token embeddings as the representation of the sub-images, summarizing their overall visual features.
Meanwhile, we use the LLM backbone to encode the question, with the hidden state of the last token in the final layer serving as its textual representation.
The relevance score \( \psi_{lm} \) is then computed as the cosine similarity between the text and image features. This method requires no additional training and leverages the modality adaptation capabilities of the multimodal projection module \( w(\cdot) \).

\smallskip
\noindent\textbf{Pretrained similarity model \( \psi_{siglip} \).}
In this approach, we leverage SigLIP \cite{siglip}, a state-of-the-art pre-trained bi-encoder model designed to measure text-image similarity. SigLIP is trained using contrastive learning to effectively assess the alignment between images and textual descriptions. We apply this model directly, without any modifications, to compute similarity scores, which are then used as relevance scores to rank the sub-images.

\smallskip
\noindent\textbf{Distant supervision \( \psi_{clip} \).}
We also introduce \( \psi_{clip} \), which requires minimal training on distant supervision data while achieving the best performance among the proposed \( \psi \) methods for \modelname. This approach fine-tunes a CLIP model to evaluate the relevance between an image and the given question.  For our primary study, we use LLaVA-1.5-7B as the base model. Since LLaVA’s supervised fine-tuning process incorporates the ScienceQA dataset—a multiple-choice VQA benchmark — we select it as the target dataset for training. 

\subsection{Training \( \psi \) with Distant Supervision}
We create the synthetic training data for \( \psi \) based on the VQA dataset.
For each VQA question-image pair \( (\mathbf{X}_v, \mathbf{X}_q) \), we first partition the high-resolution version of image \( \mathbf{X}_{v_i} \) into several non-overlapping sub-images \( \mathbf{X}_{v_i} \) and their corresponding visual tokens are \( \mathbf{H}_{v_i} \). Then, we label each sub-image that leads to the correct answer as a positive instance and each that does not as a negative instance. Since the original VQA training data are annotated with the ground-truth answer, we have the evaluator $\mathcal{P}$ and \( \mathcal{P}(f_\phi( \mathbf{H}_v, \mathbf{H}_{v_i}, \mathbf{H}_q)) \) will return the answer correctness given the low-resolution image \( \mathbf{X}_v \), the selected high-resolution sub-image \( \mathbf{X}_{v_i} \), and the question \( \mathbf{X}_q \). The positive sub-images can be defined as 
$$\mathbf{X}_{v_+}  \in \{  \mathbf{X}_{v_i} \mid \mathcal{P}(f_\phi( \mathbf{H}_v, \mathbf{H}_{v_i}, \mathbf{H}_q)) = 1 \},$$
while the negative sub-images are
$$\mathbf{X}_{v_-}  \in \{  \mathbf{X}_{v_i} \mid \mathcal{P}(f_\phi( \mathbf{H}_v, \mathbf{H}_{v_i}, \mathbf{H}_q)) = 0 \}.$$
This automatic labeling strategy yields approximately 10k question-image pairs from the ScienceQA training split, containing both positive and negative sub-images.  Using this dataset, we fine-tune a CLIP model (\texttt{openai/clip-vit-large-patch14}) with a contrastive learning approach based on margin ranking loss
\begin{equation*}
\resizebox{0.49\textwidth}{!}{
$
\mathcal{L} = \sum_{\mathbf{X}_q, \mathbf{X}_{v_+}, \mathbf{X}_{v_-}} \max(0, m + \psi(\mathbf{X}_q, \mathbf{X}_{v_-}) - \psi(\mathbf{X}_q, \mathbf{X}_{v_+})),
$}
\end{equation*}
where $m$ is the margin. This training process ensures that text-image pairs with positive sub-images receive higher relevance scores than those with negative ones. Notably, the training data is sourced from the model’s original dataset, mitigating potential data leakage concerns.



\section{Experiment}
In this section, we begin by presenting the model configurations, evaluation methods, and baseline approaches in \Cref{ssec:expSetting}. We then analyze the experimental results, discussing the effectiveness of different relevance measurements in \Cref{ssec:relevanceMesurement} and the performance improvements of \modelname over the original model in \Cref{ssec:visUnderstand}. Furthermore, we investigate the theoretical upper bound of \modelname in \Cref{ssec:optimalSelect} and introduce additional ablation studies from \Cref{ssec:majorityVote} to \Cref{ssec:ablationExtraImage}.

\begin{table*}[h!]
\centering
\setlength{\tabcolsep}{10pt}
{\fontsize{9pt}{13pt}\selectfont
\begin{tabular}{l|cccccccc}
\toprule[1.2pt]
Model& Avg. & $V^*$ & GQA & SQA & POPE & MMBench & MMStar  &SeedBench\\
\hline
\rowcolor{almond}
\multicolumn{9}{c}{\textit{Baseline Methods}} \\
LLaVA-1.5-7B     & 59.9  & 47.6  & 62.0  & 65.8  & 86.2  & 64.7 & 33.3 & 60.5 \\
$M^3$            & 60.1  & 44.0  & 61.9  & 65.5  & 87.4  & 65.9 & \textbf{35.6} & 60.6 \\
$S^2$            & 61.3  & 49.7  & 59.9  & 66.2  & 88.3  & 63.7 & 33.5 & \textbf{68.0} \\
\hline
\rowcolor{mossgreen}
\multicolumn{9}{c}{\textit{\modelname}} \\
$\psi_{lm}$                 & 59.2  & 49.2 & 59.9  & 65.6  & 84.6  & 60.5 & 31.8 & 63.0 \\
$\psi_{siglip}$             & 60.9  & \textbf{57.6}  & 60.3  & 64.5  & 86.3  & 61.6 & 32.1 & 63.7 \\
$\psi_{clip}$               & \textbf{63.2}  & 52.9  & \textbf{62.5}  & \textbf{68.3}  & \textbf{89.0}  & \textbf{72.6} & 33.5 & 63.7 \\
\bottomrule[1.2pt]
\end{tabular}
}
\caption{\modelname achieves the highest performance improvements across most tasks, surpassing all baselines in average accuracy. Notably, while baseline methods require model-specific fine-tuning, \modelname operates as a plug-and-play approach with minimal training on the top-performing task-relevance scorer, $\psi_{clip}$.}
\label{tab:main}

\end{table*}

\begin{table*}[h!]
\centering
\setlength{\tabcolsep}{10pt}
{\fontsize{9pt}{13pt}\selectfont
\begin{tabular}{l|cccccccc}
\toprule[1.2pt]
Model& Avg. & $V^*$ & GQA & SQA & POPE & MMBench & MMStar  &SeedBench\\
\rowcolor{codegray}
\multicolumn{9}{c}{\textit{Ablation study}} \\
Sub-img only        & 53.3  & 42.9  & 51.9  & 68.0  & 70.6  & 52.5  & 31.2  & 56.1 \\
Majority Vote       & 63.3  & 49.2  & 61.4  & 69.5  & 89.2  & 74.2  & 35.4  & 64.5 \\
Random              & 60.3  & 49.5  & 60.6  & 65.1  & 86.1  & 64.6  & 32.7  & 63.7 \\
$\psi_{optimal}$    &\textbf{76.0}  & \textbf{79.6} & \textbf{72.4} & \textbf{76.9}  & \textbf{95.7}  & \textbf{80.0}  & \textbf{49.7}  & \textbf{77.9} \\
\hline
LLaVA-NeXT-7B       & 63.8  & 57.1  & 64.1  & 65.7  & 87.4  & 66.7  & 35.5  & 70.0 \\
+ $\psi_{optimal}$  & 80.9  & 72.8  & 72.4  & 89.3  & 94.8  & 88.7  & 65.9  & 82.1 \\
LLaVA-1.5-13B       & 62.9  & 48.7  & 63.4  & 68.8  & 88.4  & 68.6  & 34.5  & 68.0 \\
+ $\psi_{optimal}$  & 78.6  & 82.2  & 75.6  & 78.7  & 96.5  & 83.8  & 54.3  & 79.3 \\
\bottomrule[1.2pt]
\end{tabular}
}
\caption{We perform experiments to explore the theoretical upper bound of \modelname, revealing that modern VLMs have the potential for further improvement in visual understanding tasks when provided with appropriate sub-image guidance. Additionally, ablation studies confirm that \modelname's effectiveness is not merely due to the inclusion of extra visual tokens, and that sub-images alone are insufficient for comprehensive visual understanding.}
\label{tab:upper}

\end{table*}

\subsection{Experiment Settings}\label{ssec:expSetting}
We use LLaVA-1.5 \cite{llava15} as the base VLM, with Vicuna 7B as the underlying LLM backbone. Since \modelname functions as an inference-time method, no additional training is performed. For inference, we set the temperature $\tau=0$ in all experiments to ensure reproducibility. 
To train $\psi_{clip}$, we utilize ViT-L-14 as the backbone model, fine-tuning it on the constructed training set (described in \Cref{sec:model}) for 32 epochs. The model with the lowest training loss is selected. We use a learning rate of 5e-6 and a batch size of 64.

\stitle{Evaluation} To evaluate the image understanding capabilities of our method, we test it on the following well-established benchmarks: (i) detailed understanding ($V^*$ \cite{vstar}) and (ii) VLM benchmarks for visual question answering, including POPE \cite{pope}, GQA \cite{GQA}, ScienceQA \cite{SQA}, MMBench \cite{MMBench}, MMStar \cite{MMStar}, and SeedBench \cite{seedbench}. All results are reported in terms of accuracy.

\stitle{Baselines} We compare our method against two state-of-the-art baselines built on the LLaVA architecture. Matryoshka Multimodal Model ($M^3$,~\citet{cai2024matryoshka}) employs a training strategy that dynamically compresses visual token length at different ratios during training, allowing control over visual granularity and enhancing the model’s overall visual comprehension. Scaling on Scales ($S^2$, ~\citet{shi2025we}) processes multiple cropped versions of the input image, stacks their visual features into a higher-dimensional representation, and then projects the extended visual features back into the textual feature space. By leveraging fine-grained details from sub-images, $S^2$ improves the model’s ability to understand intricate visual information. Notably, both baselines require retraining LLaVA on visual instruction data, whereas \modelname operates without additional training on the VLM itself. For a fair comparison, we use LLaVA-1.5-Vicuna-7B as the base VLM for both baselines and employ full-length visual tokens for $M^3$.

\subsection{Distant supervision yields the best performance}\label{ssec:relevanceMesurement}

The evaluation results of \modelname with three different $\psi$ methods are presented in \Cref{tab:main}.
Both $\psi_{siglip}$ and $\psi_{clip}$ outperform the base VLM by 1.0\% and 3.3\%, respectively, with $\psi_{clip}$ achieving the highest average accuracy of 63.2\%.
This suggests that $\psi_{clip}$ effectively captures task-specific signals through distant supervision, allowing it to extract the most relevant visual information from the image. At inference, the selected query-related sub-image is encoded as additional visual input tokens, providing more precise guidance for the language model.
However, \modelname with $\psi_{lm}$ performs comparably to or worse than the original model on most benchmarks. We attribute this worsened performance to the fact that the next-token embedding falls short of capturing contextual cues within the textual question, limiting its effectiveness in selecting the correct sub-image for question-answering.

\subsection{\modelname improves visual understanding}\label{ssec:visUnderstand}

As shown in \Cref{tab:main}, \modelname shows better performance when compared with other baselines. Among all compared methods, \modelname with $\psi_{clip}$ achieves the highest scores on four benchmarks, particularly excelling in MMBench, where it boosts accuracy by 7.9\%. Additionally, $\psi_{siglip}$ achieves the best performance on $V^*$, the detailed understanding benchmark. As demonstrated by \Cref{tab:upper}, $\psi_{siglip}$ even surpasses LLaVA-NeXT-7B — a model that requires additional instruction-tuning data and uses 2.5 times more visual tokens---by 0.6\%. Furthermore, \modelname with $\psi_{clip}$ improves the model’s average accuracy by 3.3\%, outperforming both $M^3$ (0.2\%) and $S^2$ (1.4\%), demonstrating its superior effectiveness in improving visual comprehension.

\modelname also offers significant efficiency advantages. Unlike $M^3$ and $S^2$, both of which require extensive retraining, \modelname is an inference-time method that operates with minimal or no additional training. Specifically, $\psi_{clip}$ is trained on approximately 3,000 data instances in under an hour using a single NVIDIA RTX 6000 ADA (48GB) GPU. In contrast, $S^2$ requires at least four hours to retrain the multimodal projection layer, while $M^3$ requires a minimum of ten hours for full instruction tuning of LLaVA on an 8$\times$NVIDIA A100 (80GB) setup.\footnote{Time estimates are based on the training specifications reported by each baseline.}



These disparities in training efficiency and benchmark performance suggest that valuable signals from VQA training data are still underutilized. The textual signal from the questions can offer significant guidance to the visual encoding process, ultimately enhancing end-task performance. This also prompts an important question that we seek to discuss next: What is the theoretical performance potential of \modelname?

\subsection{Where is the upper bound of \modelname?}\label{ssec:optimalSelect}
To explore the potential performance ceiling of \modelname, we conduct an \textit{Optimal Selection} experiment. Specifically, for a given question-image pair, the image is divided into an $n \times n$ grid\footnote{For the reported results, $n=3$.} of sub-images. Each sub-image is then paired with the original image and input into the VLM along with the textual question to generate a response. This process yields $n^2$ responses per instance. If any of these responses match the correct answer, the question is considered ``answerable'' by \modelname using optimal sub-images. We hypothesize the existence of an optimal relevance measurement, $\psi_{optimal}$, capable of selecting the best sub-image for the VLM to correctly answer all ``answerable'' questions. The performance of $\psi_{optimal}$ on each benchmark serves as the theoretical upper bound for \modelname.

As results showed in \Cref{tab:upper}, $\psi_{optimal}$ improves the original LLaVA-1.5-7B by 15.4\% on average,  which is much higher than LLaVA-NeXT-7B---the same-sized model using more training data and more visual token inputs, and LLaVA-1.5-13B that produced with the same training data but a much larger language model backbone. These observations suggest the following conclusions:  

\textbf{(1) Expanding the language model size in VLM may not be an efficient way to enhance visual understanding.} LLaVA-1.5-13B follows the same training approach and dataset as LLaVA-1.5-7B but achieves only a modest 3\% average accuracy improvement—falling short of both the 3.3\% performance gain achieved by \modelname and the theoretical 15.4\% improvement predicted by \( \psi_{optimal} \).  

This advantage of \modelname can be attributed to its strong performance on benchmarks like \( V^* \) and MMBench. In these datasets, images can be as large as 1700$\times$1500 pixels, and downscaling them to 336$\times$336 for vision encoding results in a significant loss of detail. In such cases, the vision encoder of LLaVA-1.5-13B becomes a bottleneck, whereas \modelname enhances performance by selecting the most task-relevant sub-image and providing additional detailed explanations to the VLM.

\textbf{(2) Greedily encoding all sub-areas of an image provides limited benefits.} 
The primary difference between \modelname  and LLaVA-NeXT-7B lies in their vision encoding strategies. \modelname selectively focuses on the most query-relevant sub-areas. On the contrary, LLaVa-NeXT takes a more aggressive approach by dividing the image into a 2$\times$2 grid, encoding each section separately, and appending the additional visual tokens to the encoded overview tokens. The technique of \modelname is shown to be particularly effective for tasks requiring detailed comprehension, such as \( V^* \), where it achieves a 9.5\% higher accuracy than LLaVA-1.5-7B with the same model size.  

By generating extra visual tokens, the crop-and-encode method enables higher-resolution analysis. However, its effectiveness diminishes for tasks that do not demand fine-grained details, especially with smaller images. For example, on POPE, an object detection benchmark, LLaVA-NeXT achieves only a 1.2\% accuracy improvement, as its images (approximately 333$\times$600 pixels on average) already contain sufficient information for object presence detection, rendering the additional visual tokens redundant. A similar trend is observed in SQA, a scientific question-answering dataset, where only an overview image is necessary to answer the questions---resulting in degraded performance due to the extra visual tokens.

\textbf{(3) The upper bound grows as more pretraining or larger model size.} We also conduct \( \psi_{optimal} \) experiments on both of the other LLaVA variants for comparison. In this setup, the cropped sub-images encoded by LLaVA-NeXT are replaced with a single sub-image. As shown in the bottom section of \Cref{tab:upper}, there remains a significant gap between the models' actual performance and their theoretical upper bound.  

Our observations indicate that, with expanded training data and an extended visual token length during training, the upper bound of the 7B-parameter LLaVA model increases by an average of 4.9\%. This suggests that the upper bound of \modelname is not static---\textbf{it grows as the base VLM undergoes more training with additional data}, even when the model size remains unchanged.  

Similarly, LLaVA-1.5-13B sees an average improvement of 2.6\%, with a substantial 33.5\% performance boost on \( V^* \), a detail-oriented understanding task. As discussed earlier, a larger backbone LLM benefits from a higher-dimensional latent space, which enhances modality projection. \( \psi_{optimal} \) helps overcome the vision encoder’s resolution bottleneck, making the 13B model outperform its 7B counterpart across all tasks. However, its theoretical upper bound is still 2.3\% lower than that of LLaVA-NeXT-7B.  

The discrepancy of upper bounds highlights a key insight: while LLaVA-1.5-13B improves performance by increasing the language model size, LLaVA-NeXT-7B benefits from longer visual token sequences and more training data. From this, we hypothesize that \textbf{enhancing visual understanding in VLMs is more efficiently achieved by training with more data and refining visual encoding strategies, rather than merely increasing language model size}.

\subsection{How are useful sub images distributed?}\label{ssec:majorityVote}
A key follow-up question regarding the theoretical upper bound is how accessible and densely distributed the sub-images contributing to a correct answer are. To illustrate this ``density,'' we conduct a ``Majority Vote'' experiment, where each of the \( n^2 \) sub-images is paired with the overview image to generate a response. The final answer is determined by selecting the most frequently occurring response. The results of this experiment are presented in \Cref{tab:upper}.

Overall, ``Majority Vote'' improves performance by 3.4\% on average compared to LLaVA-1.5-7B, with only a 0.1\% difference from \( \psi_{clip} \). Notably, it performs better on general VQA tasks, suggesting that in most cases, task-relevant sub-images dominate the image crops. This inherent signal enhances VLM performance but remains underutilized in current models. However, ``Majority Vote'' underperforms compared to the trained \( \psi_{clip} \) on the detailed understanding benchmark, \( V^* \), highlighting the necessity and effectiveness of selection training for high-resolution detail comprehension.

\subsection{A trade-off between image token numbers and performance}\label{ssec:tokenNums}

\begin{table}[t!]
\centering
\setlength{\tabcolsep}{10pt}
{\fontsize{9pt}{13pt}\selectfont
\begin{tabular}{l|ccc}
\toprule[1.2pt]

Method           & \#img tokens& \#Params. & Avg. Acc \\
\hline
LLaVA-1.5       & 576           & 7B  &   59.9   \\
$M^3$           & 576           & 7B  &   60.1   \\
$S^2$           & 576           & 7B  &   61.3   \\
LLaVA-1.5       & 576           & 13B &   62.9   \\
\modelname      & 576$\times$2  & 7B  &   63.2   \\
LLaVA-NeXT      & 576$\times$5  & 7B  &   63.8   \\
\bottomrule[1.2pt]
\end{tabular}
}
\caption{\modelname delivers performance on par with the best baseline while requiring significantly less training and 60\% fewer visual tokens, achieving the optimal balance between training cost, inference efficiency, and overall accuracy. Additionally, \modelname outperforms a VLM with a substantially larger language model, highlighting the superiority of task-guided visual selection over simply increasing language model size.}
\label{tab:numTok}

\end{table}

We analyze the trade-off between efficiency and performance across different methods. Based on experiments varying the number \( k \) of top-relevant sub-images used by \modelname, the best performance is consistently achieved at \( k=1 \), effectively doubling the number of visual tokens fed into the VLM. We present a comparison of the number of processed image tokens, total VLM parameters, and overall performance on visual understanding tasks, as detailed in \Cref{tab:numTok}. Compared to the best-performing LLaVA-NeXT, \modelname-equipped LLaVA-1.5 achieves similar results with the same model size while requiring 60\% fewer visual tokens. Notably, \modelname is a plug-and-play method that does not require any additional training for the VLM. Furthermore, compared to LLaVA-1.5-13B, which uses nearly twice the language model size, \modelname outperforms it, particularly on fine-grained detail understanding tasks. We conclude that \modelname effectively encodes fine-grained visual information by selectively pruning visual tokens based on the semantic relevance of the task query.

\subsection{Is the sub-image itself enough?}
To highlight the importance of pairing the overview image, we conduct the \textit{Sub-img only} experiment. In this setup, we remove the visual tokens from the original image and provide only the encoded sub-image to the language model for evaluation. On average, performance drops by 6.6\%, indicating that the overview image is still necessary to the effectiveness of \modelname.

\subsection{Extra visual tokens do not guarantee improvement}\label{ssec:ablationExtraImage}
Another potential concern is whether the performance improvement of \modelname stems from the additional computation required by extra visual tokens. Since the language model processes a longer input sequence, 
we wonder if the performance is improved at the cost of increased FLOPs and computation time.
To address this concern and ensure a fair comparison, we conduct an experiment where a randomly selected sub-image is paired with the original image as visual input. This method is reported as ``Random'' in \Cref{tab:upper}. We repeat this process 32 times to calculate the average performance.

As shown in the table, the ``Random'' approach slightly outperforms the original LLaVA-1.5-7B. However, a clear improvement of 1.9\% on $V^*$ and 3.2\% on SeedBench is observed. Based on human analysis, the improvement on $V^*$ is likely due to the resolution bottleneck of the vision encoder---a randomly chosen sub-image may carry task-relevant, zoomed-in details that help answer the question correctly. As for the performance gain on SeedBench, it primarily comes from the ``instance attribution'' sub-task, which assesses an item's properties in an image. We hypothesize that a randomly selected sub-image may sometimes capture and emphasize such information, making it more salient to the language model and enhancing performance.




\section{Related Work}

\vspace{1ex} \noindent \textbf{Vision Language Models}. LLMs have exhibited impressive performance on complex tasks such as mathematical reasoning and coding~\citep{dubey2024llama,jaech2024openai}. With vision encoders incorporated for visual information perception and projectors or adaptors for modality alignment, VLMs are able to perform visual question answering~\citep{zhang2024internlm,wang2024qwen2}. 

Typically, VLMs have a pre-trained vision encoder to represent visual features, a pre-trained LLM to understand user instructions and provide responses, and a vision-language cross-modal connector
to align the vision encoder outputs to the language models. Training VLMs includes two stages: Image caption data~\citep{schuhmann2022laion,singh2022flava} and interleaved image-text data~\citep{laurenccon2024obelics,zhu2024multimodal} are used to align vision and language modality in the pre-training stage, while learning from instruction following data~\citep{llava10,zhang2023llavar} enables models to understand instructions including visual content.
Given an instruction-tuned model, this work proposes a plug-and-play way to integrate textual signals into visual encoding process, which further improves visual understanding without additional training.

\vspace{1ex} \noindent \textbf{Cross-modality information grounding}. To map representations from visual encoders to the LLM token space, MLP projectors are highly effective and more lightweight compared with prior connectors such as Resamplers~\citep{alayrac2022flamingo} and Q-Formers~\citep{bai2023qwen,dai2023instructblip}. However, understanding high-resolution images for applications such as OCR and chart analysis presents significant challenges for MLP projectors. The visual token count grows quadratically with image resolution~\citep{tong2024cambrian}, when directly adopting high-resolution visual encoders~\citep{li2024mini,lv2023kosmos} or utilizating patchification (i.e., cropping original images into patches and concatenating visual embeddings)~\citep{li2023otterhd,dong2024internlm,li2024monkey}. Recently, training strategies, such as Matryoshka Multimodal Model ($M^3$,~\citet{cai2024matryoshka}) and Scaling on Scales ($S^2$, ~\citet{shi2025we}), have been proposed to further fine-tune VLMs on instruction data for high-resolution visual information understanding. \citep{visualcot} also proposed to finetune VLMs to draw the bounding box of the question-related area, leading to better detail understanding.

Notably, our proposed \modelname is more efficient without additional training while more effective with textual signals incorporated into visual encoding.
\section{Conclusion}
In this work, we introduce \modelname, a plug-and-play approach that enhances VLMs' visual understanding through semantic-guided visual selection. Integrating \modelname significantly improves VLMs' ability to comprehend fine-grained visual details while requiring minimal training for the relevance scorer. Furthermore, we demonstrate that \modelname has a remarkably high theoretical upper bound, with its potential increasing as the base VLM is trained on more data.

Our work paves the way for future research on modality fusion, exploring whether textual signals can be leveraged to distill task-relevant visual information, ultimately leading to a more efficient and effective multimodal learning system. Additionally, our upper-bound experiment on \modelname highlights the untapped potential of current VLMs, which is constrained by existing learning paradigms. Enhancing relevance measurement or involving semantic-guided visual selection during VLM training could help unlock these latent capabilities.

\section*{Impact Statement}
Our work investigates the critical challenge in fine-grained visual reasoning, emphasizing its social impact, including better perception of high-resolution images and reliable responses for high-stakes applications. The adoption of the open-resource MLLM and evaluation on open benchmarks ensure data transparency and result reproducibility. Our proposed strategy aims for effective and efficient visual understanding, which aligns with the goal of trustworthy and sustainable AI.



\nocite{langley00}

\bibliography{custom}

\begin{thebibliography}{38}
\providecommand{\natexlab}[1]{#1}
\providecommand{\url}[1]{\texttt{#1}}
\expandafter\ifx\csname urlstyle\endcsname\relax
  \providecommand{\doi}[1]{doi: #1}\else
  \providecommand{\doi}{doi: \begingroup \urlstyle{rm}\Url}\fi

\bibitem[Alayrac et~al.(2022{\natexlab{a}})Alayrac, Donahue, Luc, Miech, Barr, Hasson, Lenc, Mensch, Millicah, Reynolds, Ring, Rutherford, Cabi, Han, Gong, Samangooei, Monteiro, Menick, Borgeaud, Brock, Nematzadeh, Sharifzadeh, Binkowski, Barreira, Vinyals, Zisserman, and Simonyan]{Flamingo}
Alayrac, J.-B., Donahue, J., Luc, P., Miech, A., Barr, I., Hasson, Y., Lenc, K., Mensch, A., Millicah, K., Reynolds, M., Ring, R., Rutherford, E., Cabi, S., Han, T., Gong, Z., Samangooei, S., Monteiro, M., Menick, J., Borgeaud, S., Brock, A., Nematzadeh, A., Sharifzadeh, S., Binkowski, M., Barreira, R., Vinyals, O., Zisserman, A., and Simonyan, K.
\newblock Flamingo: a visual language model for few-shot learning.
\newblock NIPS '22, Red Hook, NY, USA, 2022{\natexlab{a}}. Curran Associates Inc.
\newblock ISBN 9781713871088.

\bibitem[Alayrac et~al.(2022{\natexlab{b}})Alayrac, Donahue, Luc, Miech, Barr, Hasson, Lenc, Mensch, Millican, Reynolds, et~al.]{alayrac2022flamingo}
Alayrac, J.-B., Donahue, J., Luc, P., Miech, A., Barr, I., Hasson, Y., Lenc, K., Mensch, A., Millican, K., Reynolds, M., et~al.
\newblock Flamingo: a visual language model for few-shot learning.
\newblock \emph{Advances in neural information processing systems}, 35:\penalty0 23716--23736, 2022{\natexlab{b}}.

\bibitem[Bai et~al.(2023)Bai, Bai, Yang, Wang, Tan, Wang, Lin, Zhou, and Zhou]{bai2023qwen}
Bai, J., Bai, S., Yang, S., Wang, S., Tan, S., Wang, P., Lin, J., Zhou, C., and Zhou, J.
\newblock Qwen-vl: A versatile vision-language model for understanding, localization, text reading, and beyond.
\newblock \emph{arXiv preprint arXiv:2308.12966}, 1\penalty0 (2):\penalty0 3, 2023.

\bibitem[Cai et~al.(2024)Cai, Yang, Gao, and Lee]{cai2024matryoshka}
Cai, M., Yang, J., Gao, J., and Lee, Y.~J.
\newblock Matryoshka multimodal models.
\newblock \emph{arXiv preprint arXiv:2405.17430}, 2024.

\bibitem[Chen et~al.(2023)Chen, Zhang, Zeng, Zhang, Zhu, and Zhao]{chen2023shikra}
Chen, K., Zhang, Z., Zeng, W., Zhang, R., Zhu, F., and Zhao, R.
\newblock Shikra: Unleashing multimodal llm's referential dialogue magic.
\newblock \emph{arXiv preprint arXiv:2306.15195}, 2023.

\bibitem[Chen et~al.(2024)Chen, Li, Dong, Zhang, Zang, Chen, Duan, Wang, Qiao, Lin, et~al.]{MMStar}
Chen, L., Li, J., Dong, X., Zhang, P., Zang, Y., Chen, Z., Duan, H., Wang, J., Qiao, Y., Lin, D., et~al.
\newblock Are we on the right way for evaluating large vision-language models?
\newblock \emph{arXiv preprint arXiv:2403.20330}, 2024.

\bibitem[Dai et~al.(2023)Dai, Li, Li, Tiong, Zhao, Wang, Li, Fung, and Hoi]{dai2023instructblip}
Dai, W., Li, J., Li, D., Tiong, A., Zhao, J., Wang, W., Li, B., Fung, P., and Hoi, S.
\newblock Instructblip: Towards general-purpose vision-language models with instruction tuning. arxiv 2023.
\newblock \emph{arXiv preprint arXiv:2305.06500}, 2, 2023.

\bibitem[Dong et~al.(2024)Dong, Zhang, Zang, Cao, Wang, Ouyang, Zhang, Duan, Zhang, Li, et~al.]{dong2024internlm}
Dong, X., Zhang, P., Zang, Y., Cao, Y., Wang, B., Ouyang, L., Zhang, S., Duan, H., Zhang, W., Li, Y., et~al.
\newblock Internlm-xcomposer2-4khd: A pioneering large vision-language model handling resolutions from 336 pixels to 4k hd.
\newblock \emph{arXiv preprint arXiv:2404.06512}, 2024.

\bibitem[Dosovitskiy(2020)]{vit}
Dosovitskiy, A.
\newblock An image is worth 16x16 words: Transformers for image recognition at scale.
\newblock \emph{arXiv preprint arXiv:2010.11929}, 2020.

\bibitem[Dubey et~al.(2024)Dubey, Jauhri, Pandey, Kadian, Al-Dahle, Letman, Mathur, Schelten, Yang, Fan, et~al.]{dubey2024llama}
Dubey, A., Jauhri, A., Pandey, A., Kadian, A., Al-Dahle, A., Letman, A., Mathur, A., Schelten, A., Yang, A., Fan, A., et~al.
\newblock The llama 3 herd of models.
\newblock \emph{arXiv preprint arXiv:2407.21783}, 2024.

\bibitem[He et~al.(2024)He, Liu, Wu, Yuan, Wang, Huang, and Zhao]{he2024efficient}
He, M., Liu, Y., Wu, B., Yuan, J., Wang, Y., Huang, T., and Zhao, B.
\newblock Efficient multimodal learning from data-centric perspective.
\newblock \emph{arXiv preprint arXiv:2402.11530}, 2024.

\bibitem[Hudson \& Manning(2019)Hudson and Manning]{GQA}
Hudson, D.~A. and Manning, C.~D.
\newblock Gqa: A new dataset for real-world visual reasoning and compositional question answering.
\newblock In \emph{Proceedings of the IEEE/CVF Conference on Computer Vision and Pattern Recognition (CVPR)}, June 2019.

\bibitem[Jaech et~al.(2024)Jaech, Kalai, Lerer, Richardson, El-Kishky, Low, Helyar, Madry, Beutel, Carney, et~al.]{jaech2024openai}
Jaech, A., Kalai, A., Lerer, A., Richardson, A., El-Kishky, A., Low, A., Helyar, A., Madry, A., Beutel, A., Carney, A., et~al.
\newblock Openai o1 system card.
\newblock \emph{arXiv preprint arXiv:2412.16720}, 2024.

\bibitem[Lauren{\c{c}}on et~al.(2024)Lauren{\c{c}}on, Saulnier, Tronchon, Bekman, Singh, Lozhkov, Wang, Karamcheti, Rush, Kiela, et~al.]{laurenccon2024obelics}
Lauren{\c{c}}on, H., Saulnier, L., Tronchon, L., Bekman, S., Singh, A., Lozhkov, A., Wang, T., Karamcheti, S., Rush, A., Kiela, D., et~al.
\newblock Obelics: An open web-scale filtered dataset of interleaved image-text documents.
\newblock \emph{Advances in Neural Information Processing Systems}, 36, 2024.

\bibitem[Li et~al.(2023{\natexlab{a}})Li, Wang, Wang, Ge, Ge, and Shan]{seedbench}
Li, B., Wang, R., Wang, G., Ge, Y., Ge, Y., and Shan, Y.
\newblock Seed-bench: Benchmarking multimodal llms with generative comprehension.
\newblock \emph{arXiv preprint arXiv:2307.16125}, 2023{\natexlab{a}}.

\bibitem[Li et~al.(2023{\natexlab{b}})Li, Zhang, Yang, Zhang, Pu, and Liu]{li2023otterhd}
Li, B., Zhang, P., Yang, J., Zhang, Y., Pu, F., and Liu, Z.
\newblock Otterhd: A high-resolution multi-modality model.
\newblock \emph{arXiv preprint arXiv:2311.04219}, 2023{\natexlab{b}}.

\bibitem[Li et~al.(2023{\natexlab{c}})Li, Du, Zhou, Wang, Zhao, and Wen]{pope}
Li, Y., Du, Y., Zhou, K., Wang, J., Zhao, X., and Wen, J.-R.
\newblock Evaluating object hallucination in large vision-language models.
\newblock In Bouamor, H., Pino, J., and Bali, K. (eds.), \emph{Proceedings of the 2023 Conference on Empirical Methods in Natural Language Processing}, pp.\  292--305, Singapore, December 2023{\natexlab{c}}. Association for Computational Linguistics.
\newblock \doi{10.18653/v1/2023.emnlp-main.20}.
\newblock URL \url{https://aclanthology.org/2023.emnlp-main.20/}.

\bibitem[Li et~al.(2024{\natexlab{a}})Li, Zhang, Wang, Zhong, Chen, Chu, Liu, and Jia]{li2024mini}
Li, Y., Zhang, Y., Wang, C., Zhong, Z., Chen, Y., Chu, R., Liu, S., and Jia, J.
\newblock Mini-gemini: Mining the potential of multi-modality vision language models.
\newblock \emph{arXiv preprint arXiv:2403.18814}, 2024{\natexlab{a}}.

\bibitem[Li et~al.(2024{\natexlab{b}})Li, Yang, Liu, Ma, Zhang, Yang, Sun, Liu, and Bai]{li2024monkey}
Li, Z., Yang, B., Liu, Q., Ma, Z., Zhang, S., Yang, J., Sun, Y., Liu, Y., and Bai, X.
\newblock Monkey: Image resolution and text label are important things for large multi-modal models.
\newblock In \emph{Proceedings of the IEEE/CVF Conference on Computer Vision and Pattern Recognition}, pp.\  26763--26773, 2024{\natexlab{b}}.

\bibitem[Liu et~al.(2023)Liu, Li, Wu, and Lee]{llava10}
Liu, H., Li, C., Wu, Q., and Lee, Y.~J.
\newblock Visual instruction tuning.
\newblock In Oh, A., Naumann, T., Globerson, A., Saenko, K., Hardt, M., and Levine, S. (eds.), \emph{Advances in Neural Information Processing Systems}, volume~36, pp.\  34892--34916. Curran Associates, Inc., 2023.

\bibitem[Liu et~al.(2024{\natexlab{a}})Liu, Li, Li, and Lee]{llava15}
Liu, H., Li, C., Li, Y., and Lee, Y.~J.
\newblock Improved baselines with visual instruction tuning.
\newblock In \emph{Proceedings of the IEEE/CVF Conference on Computer Vision and Pattern Recognition}, pp.\  26296--26306, 2024{\natexlab{a}}.

\bibitem[Liu et~al.(2024{\natexlab{b}})Liu, Li, Li, Li, Zhang, Shen, and Lee]{llavanext}
Liu, H., Li, C., Li, Y., Li, B., Zhang, Y., Shen, S., and Lee, Y.~J.
\newblock Llava-next: Improved reasoning, ocr, and world knowledge, January 2024{\natexlab{b}}.
\newblock URL \url{https://llava-vl.github.io/blog/2024-01-30-llava-next/}.

\bibitem[Liu et~al.(2024{\natexlab{c}})Liu, Duan, Zhang, Li, Zhang, Zhao, Yuan, Wang, He, Liu, et~al.]{MMBench}
Liu, Y., Duan, H., Zhang, Y., Li, B., Zhang, S., Zhao, W., Yuan, Y., Wang, J., He, C., Liu, Z., et~al.
\newblock Mmbench: Is your multi-modal model an all-around player?
\newblock In \emph{European conference on computer vision}, pp.\  216--233. Springer, 2024{\natexlab{c}}.

\bibitem[Lu et~al.(2022)Lu, Mishra, Xia, Qiu, Chang, Zhu, Tafjord, Clark, and Kalyan]{SQA}
Lu, P., Mishra, S., Xia, T., Qiu, L., Chang, K.-W., Zhu, S.-C., Tafjord, O., Clark, P., and Kalyan, A.
\newblock Learn to explain: Multimodal reasoning via thought chains for science question answering.
\newblock In \emph{The 36th Conference on Neural Information Processing Systems (NeurIPS)}, 2022.

\bibitem[Lv et~al.(2023)Lv, Huang, Chen, Zhao, Jia, Cui, Ma, Chang, Huang, Wang, et~al.]{lv2023kosmos}
Lv, T., Huang, Y., Chen, J., Zhao, Y., Jia, Y., Cui, L., Ma, S., Chang, Y., Huang, S., Wang, W., et~al.
\newblock Kosmos-2.5: A multimodal literate model.
\newblock \emph{arXiv preprint arXiv:2309.11419}, 2023.

\bibitem[Radford et~al.(2021)Radford, Kim, Hallacy, Ramesh, Goh, Agarwal, Sastry, Askell, Mishkin, Clark, et~al.]{clip}
Radford, A., Kim, J.~W., Hallacy, C., Ramesh, A., Goh, G., Agarwal, S., Sastry, G., Askell, A., Mishkin, P., Clark, J., et~al.
\newblock Learning transferable visual models from natural language supervision.
\newblock In \emph{International conference on machine learning}, pp.\  8748--8763. PMLR, 2021.

\bibitem[Schuhmann et~al.(2022)Schuhmann, Beaumont, Vencu, Gordon, Wightman, Cherti, Coombes, Katta, Mullis, Wortsman, et~al.]{schuhmann2022laion}
Schuhmann, C., Beaumont, R., Vencu, R., Gordon, C., Wightman, R., Cherti, M., Coombes, T., Katta, A., Mullis, C., Wortsman, M., et~al.
\newblock Laion-5b: An open large-scale dataset for training next generation image-text models.
\newblock \emph{Advances in Neural Information Processing Systems}, 35:\penalty0 25278--25294, 2022.

\bibitem[Shao et~al.(2024)Shao, Qian, Xiao, Song, Zong, Wang, Liu, and Li]{visualcot}
Shao, H., Qian, S., Xiao, H., Song, G., Zong, Z., Wang, L., Liu, Y., and Li, H.
\newblock Visual cot: Unleashing chain-of-thought reasoning in multi-modal language models, 2024.

\bibitem[Shi et~al.(2025)Shi, Wu, Mao, Wang, and Darrell]{shi2025we}
Shi, B., Wu, Z., Mao, M., Wang, X., and Darrell, T.
\newblock When do we not need larger vision models?
\newblock In \emph{European Conference on Computer Vision}, pp.\  444--462. Springer, 2025.

\bibitem[Singh et~al.(2022)Singh, Hu, Goswami, Couairon, Galuba, Rohrbach, and Kiela]{singh2022flava}
Singh, A., Hu, R., Goswami, V., Couairon, G., Galuba, W., Rohrbach, M., and Kiela, D.
\newblock Flava: A foundational language and vision alignment model.
\newblock In \emph{Proceedings of the IEEE/CVF Conference on Computer Vision and Pattern Recognition}, pp.\  15638--15650, 2022.

\bibitem[Thapa et~al.(2024)Thapa, Chen, Covert, Chalamala, Athiwaratkun, Song, and Zou]{dragonfly}
Thapa, R., Chen, K., Covert, I., Chalamala, R., Athiwaratkun, B., Song, S.~L., and Zou, J.
\newblock Dragonfly: Multi-resolution zoom-in encoding enhances vision-language models, 2024.

\bibitem[Tong et~al.(2024)Tong, Brown, Wu, Woo, Middepogu, Akula, Yang, Yang, Iyer, Pan, et~al.]{tong2024cambrian}
Tong, S., Brown, E., Wu, P., Woo, S., Middepogu, M., Akula, S.~C., Yang, J., Yang, S., Iyer, A., Pan, X., et~al.
\newblock Cambrian-1: A fully open, vision-centric exploration of multimodal llms.
\newblock \emph{arXiv preprint arXiv:2406.16860}, 2024.

\bibitem[Wang et~al.(2024)Wang, Bai, Tan, Wang, Fan, Bai, Chen, Liu, Wang, Ge, et~al.]{wang2024qwen2}
Wang, P., Bai, S., Tan, S., Wang, S., Fan, Z., Bai, J., Chen, K., Liu, X., Wang, J., Ge, W., et~al.
\newblock Qwen2-vl: Enhancing vision-language model's perception of the world at any resolution.
\newblock \emph{arXiv preprint arXiv:2409.12191}, 2024.

\bibitem[Wu \& Xie(2023)Wu and Xie]{vstar}
Wu, P. and Xie, S.
\newblock V*: Guided visual search as a core mechanism in multimodal llms.
\newblock \emph{arXiv preprint arXiv:2312.14135}, 2023.

\bibitem[Zhai et~al.(2023)Zhai, Mustafa, Kolesnikov, and Beyer]{siglip}
Zhai, X., Mustafa, B., Kolesnikov, A., and Beyer, L.
\newblock Sigmoid loss for language image pre-training, 2023.

\bibitem[Zhang et~al.(2024)Zhang, Dong, Zang, Cao, Qian, Chen, Guo, Duan, Wang, Ouyang, et~al.]{zhang2024internlm}
Zhang, P., Dong, X., Zang, Y., Cao, Y., Qian, R., Chen, L., Guo, Q., Duan, H., Wang, B., Ouyang, L., et~al.
\newblock Internlm-xcomposer-2.5: A versatile large vision language model supporting long-contextual input and output.
\newblock \emph{arXiv preprint arXiv:2407.03320}, 2024.

\bibitem[Zhang et~al.(2023)Zhang, Zhang, Gu, Zhou, Lipka, Yang, and Sun]{zhang2023llavar}
Zhang, Y., Zhang, R., Gu, J., Zhou, Y., Lipka, N., Yang, D., and Sun, T.
\newblock Llavar: Enhanced visual instruction tuning for text-rich image understanding.
\newblock \emph{arXiv preprint arXiv:2306.17107}, 2023.

\bibitem[Zhu et~al.(2024)Zhu, Hessel, Awadalla, Gadre, Dodge, Fang, Yu, Schmidt, Wang, and Choi]{zhu2024multimodal}
Zhu, W., Hessel, J., Awadalla, A., Gadre, S.~Y., Dodge, J., Fang, A., Yu, Y., Schmidt, L., Wang, W.~Y., and Choi, Y.
\newblock Multimodal c4: An open, billion-scale corpus of images interleaved with text.
\newblock \emph{Advances in Neural Information Processing Systems}, 36, 2024.

\end{thebibliography}
\bibliographystyle{icml2025}




\end{document}